ROBOMECH Journal

# Evaluation of two complementary modeling approaches for fiber-reinforced soft actuators

Soheil Habibian[1,2]* , Benjamin B. Wheatley[3], Suehye Bae[3], Joon Shin[3] and Keith W. Buffinton[3]

## Abstract

Although robots are increasingly found in a wide range of applications, their use in proximity to humans is still fraught with challenges, primarily due to safety concerns. Roboticists have been seeking to address this situation in recent years through the use of soft robots. Unfortunately, identifying appropriate models for the complete analysis and investigation of soft robots for design and control purposes can be problematic. This paper seeks to address this challenge by proposing two complementary modeling techniques for a particular type of soft robotic actuator known as a Fiber-Reinforced Elastomeric Enclosure (FREE). We propose that researchers can leverage multiple models to fill gaps in the understanding of the behavior of soft robots. We present and evaluate both a dynamic, lumped-parameter model and a finite element model to extend understanding of the practicability of FREEs in soft robotic applications. The results of experimental simulations using a lumped-parameter model show that at low pressures FREE winding angle and radius change no more than 2%. This observation provided confidence that a linearized, dynamic, lumped-mass model could be successfully used for FREE controller development. Results with the lumped-parameter model demonstrate that it predicts the actual rotational motion of a FREE with at most 4% error when a closed-loop controller is embedded in the system. Additionally, finite element analysis was used to study FREE design parameters as well as the workspace achieved with a module comprised of multiple FREEs. Our finite element results indicate that variations in the material properties of the elastic enclosure of a FREE are more significant than variations in fiber properties (primarily because the fibers are essentially inextensible in comparison to the elastic enclosure). Our finite element analysis confirms the results obtained by previous researchers for the impact of variations in winding angle on FREE rotation, and we extend these results to include an analysis of the effect of winding angle on FREE force and moment generation. Finally, finite element results show that a 30° difference in winding angle dramatically alters the shape of the workspace generated by four FREEs assembled into a module. Concludingly, comments are made about the relative advantages and limitations of lumped-parameter and finite element models of FREEs and FREE modules in providing useful insights into their behavior.

## Introduction

As humans seek more efficient and convenient lifestyles through new technologies, such as robotics, the use of traditional materials can limit the incorporation of technological advances into daily life. To date, robots have proven to be practical in industry, where they are used for dirty, dull, dangerous, domestic, and dexterous tasks. On the other hand, there is a high level of uncertainty around human-robot interactions [1]. Roboticists attempt to alleviate problems that humans face when working in proximity to robots. Isolating robots in protective cages is used in traditional robotics systems [2], a game-theoretic approach to constrain robot agents to stay within their territory while completing the tasks [3], or employing active robot learning methods to perceive the human's preference

*Correspondence: habibian@vt.edu
[1] Was with Department of Mechanical Engineering, Bucknell University, Lewisburg, PA, USA
Full list of author information is available at the end of the article

[4] are examples of several of the approaches to make robots behave safely and desirably. This motivated the sub-domain of the soft robotics field in which soft materials are employed to increase flexibility for robot developers [14]. Soft robots that often mimic the behavior of animals and plants in nature [15] exceed the capabilities of traditional robots and make them favorable for various robotic applications. As opposed to traditional rigid robots [16], soft robots can adapt their configuration to unpredictable environments [17], benefit from high degrees of freedom [18], are safer when they are in contact with humans because of their soft materials [9], and have cheaper and more accessible fabrication processes [19].

In the last several years, soft robots (Fig. 1) have appeared as manipulators appropriate for sophisticated motions [20], artificial skins in biomechanical applications [21], autonomous mobile robots in constrained environments [17, 22], delicate grippers [23], and even underwater robots [24, 25]. All of the existing work in this field emphasizes that having structurally soft components profoundly decreases the destructive interaction between humans, machines, and environments. Consequently, soft robots have a significant role to play where robots interact closely with the environment and humans.

The essential components of soft robots are soft actuators that typically are based on electrical [26], thermal [27], pneumatic [28], or hydraulic actuation [29]. Soft robots, regardless of their physical structures and applications, share one common characteristic: modeling and controlling soft robotic systems is still challenging for roboticists due to their large number of degrees of freedom and inherent nonlinearity. However, researchers have employed different approaches to modeling soft robots. From pure kinematic models [30] to linearized analytical models [28], to quasi-static models [31], to finite element models [32], to data-driven models [33, 34], and physics-based static models [35], all of these approaches are justified in certain soft robotic systems. Depending on the type of actuator and its application, one approach may be favorable to another; and of course, all models have shortcomings in specific circumstances and are impractical under certain assumptions [36]. Incorporating physical constraints into systems has also been used to make modeling more convenient and accurate [37].

For the scope of this work, we present results based on the utilization of a combination of two separate models for a sub-category of soft actuators. Our study aims to characterize the behavior of Fiber-Reinforced

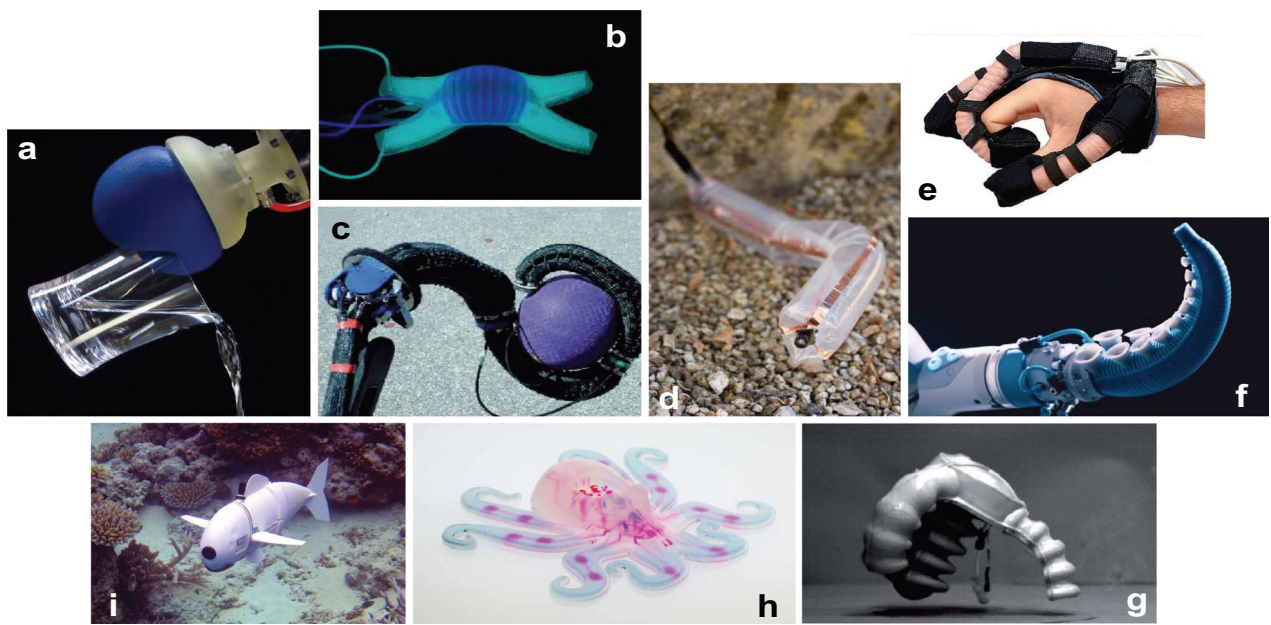

**Fig. 1** Soft robots inspired by nature demonstrating superior performance over traditional robots. **a** Jamming-based soft gripper picking up a wide range of objects [5], **b** Soft quadruped robot glowing in the dark using chemiluminescence [6], **c** Soft-continuum manipulator grasping a sphere during field trials [7], **d** Soft growing robot navigating by exploiting contact with terrain [8], **e** Soft rehabilitation glove assisting humans with functional grasp pathologies [9], **f** Tapered soft gripper inspired by octopus arms providing high dexterity and flexibility [10], **g** Untethered soft-bodied robot executing agile locomotion [11], **h** Soft, autonomous robot performing untethered operation [12], **i** Soft robotic fish exploring underwater along coral reefs in the Pacific Ocean [13]. Images have been obtained from publicly available resources and their intellectual property belongs to the authors, sources, and publishers of the cited references

Elastomeric Enclosure, called a FREE (Fig. 2). FREEs, as a composite pneumatic actuator, consist of an elastomer tube (latex) as the matrix, and wound strings (cotton) as the fiber reinforcement. Intuitively, the latex tube binds and shapes the fibers, while the cotton fibers contribute to the strength and stiffness of the actuator. Pressurized FREEs generate principal motions that include axial rotation and displacement, which serve as building blocks of more sophisticated motions when they are used in assemblies.

The choice of geometric parameters or materials used in FREEs changes their behavior. However finding appropriate parameters is a difficult task that raises critical questions: What combination of parameters best contributes to the desired behavior? How do FREEs behave if assembled in a module? Is the actuator even controllable? FREEs have been studied with various modeling approaches that each address their behavior under certain assumptions and broadly evaluate the practicality of the modeling technique. A kinematic model was introduced in [38] using the geometric relationship between fibers and fluid forces. A later work [30] found that fiber alignment on the tube significantly affects the motion, which may cause a condition called a locked manifold (i.e., the FREE does not deform with inflation). Based on these studies, FREEs' geometry have been simplified to create a quasi-static model [39] to perform open-loop control and estimate design parameters. Additionally, the nonlinear characteristics of FREEs were computationally studied [40] to illustrate the relationship between input pressure and resulting forces and deformations through a continuum model. A significant step forward was research on parallel combinations of FREEs [35, 38] assembled into a module. Each of these studies has contributed significantly to characterizing FREEs' behavior but illustrates that no one model sufficiently characterizes FREE behavior, particularly when multiple FREEs are coupled into a module. We believe that designers can leverage complementary modeling approaches to fill the gaps in the understanding of soft robotic actuators. Our goal is to characterize and control FREEs' behavior based on both a lumped-parameter dynamic model and a finite element material model.

Overall, we make the following contributions: (1) a dynamic model that characterizes the behavior of a single FREE, (2) a model-driven controller simulation and experiment to visualize single FREEs' time-dependent response in following a desired rotation, (3) a finite element model of FREEs in single and module configurations as an additional design and verification tool, (4) an exploration of the locations of points within the workspace of a module of multiple FREEs, and (5) an evaluation of the application of both modeling techniques for fiber-reinforced soft robotic actuators.

The remainder of the paper is organized as follows: the Lumped Parameter Model section presents the equations of motions of a single FREE and their relationship to the input pressure. The Controller Design section provides insight into the PID control of a single FREE. The Finite Element Model section describes the formulation and numerical modeling of a finite element model of FREEs. The Significance of Winding Angle section presents the results of a parametric study using the finite element model and discusses the role of fiber winding angles in FREEs' behavior. The FREE Module section analyzes the workspace of multiple FREEs in a module using a finite element model. Next, we present an Experimental Demonstration of FREEs in a real-life application. And finally, the Conclusion section, presents conclusions about the broader implications of the proposed modeling techniques and how they supplement each other. The overreaching goal is to demonstrate the capabilities of FREEs as an actuator for soft robots performing typical daily tasks, by exploiting the created tools and controller.

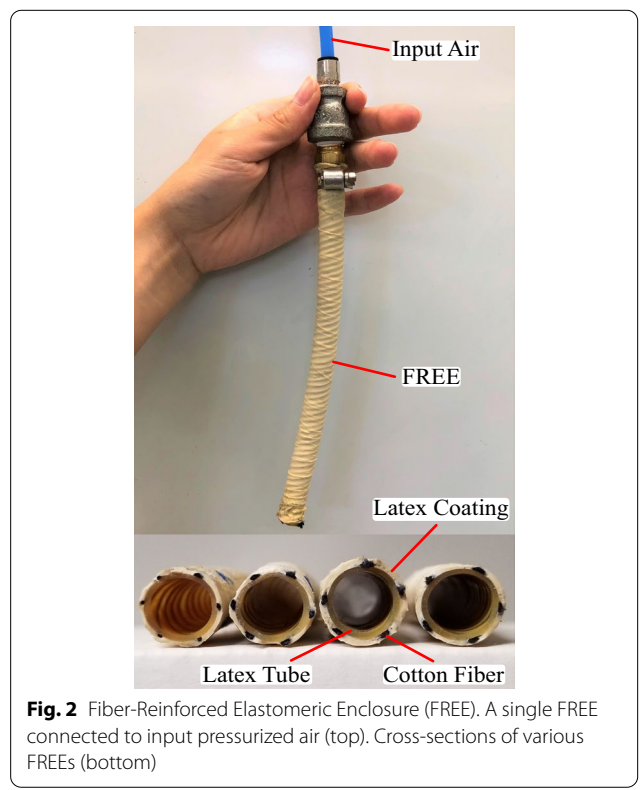

**Fig. 2** Fiber-Reinforced Elastomeric Enclosure (FREE). A single FREE connected to input pressurized air (top). Cross-sections of various FREEs (bottom)

## Lumped parameter model

This section formulates the equations describing the principal motions of a single FREE: rotation and elongation. Establishing a relationship between these motions and input pressure is the primary interest. This problem has already been tackled in prior work [39] and the static behavior of FREEs based on design parameters was studied. In this paper, we emphasize studying the dynamics of the system and explore controlled time-dependent responses. We formulate a dynamic lumped parameter model that includes mass, moment of inertia, stiffness constants, and damping coefficients. To simplify the analysis, the FREE is assumed to be a uniform thin-walled tube, wound with inextensible fibers with a constant cross-section that are perfectly adhered to the outer surface of the tube (Fig. 3). Using these simplifying assumptions enables the equations of motion to be derived based on fiber tension, elastomer force, elastomer moment, and the input pressure.

Figure 3 illustrates the geometric parameters of a FREE when it is fixed at one end. The FREE has a radius $r$, fiber winding angle $\gamma$, and length $l$ at an applied pneumatic pressure $P$. We assume that these parameters are variable depending on rotation angle $\phi$ and displacement $s$ at the free end. Here lowercase notation is used for parameters after pressurization, and uppercase denotes the initial values. The geometrical relationships between these parameters is addressed in detail in [41].

The elastomer force $F_e$ and moment $M_e$ are characterized by two simplifications: decoupling and linearity. Based on the decoupling assumption, the axial elastomer force $F_e$ depends only on the axial motion $s$, and the elastomer moment $M_e$ is purely dependent on the rotation angle $\phi$. We make these assumptions to facilitate the modeling procedure, and as shown later, to analyze the system more easily.

The elastomer is a nonlinear hyper-elastic material (see Finite Element Model section), which makes relating displacement to force challenging. However, this non-linearity arises at large strains so we assume a linear relationship for small strains at low pressures. Hence, the force $F_e$ and moment $M_e$ created by the elastomer are modeled as linear functions of the free end extension $s$ and rotation $\phi$ and their derivatives:

$$F_e = -k_e s - c_e \dot{s} \tag{1}$$

$$M_e = -k_t \phi - c_t \dot{\phi} \tag{2}$$

where $k_e$ and $k_t$ are the linear and torsional stiffnesses, and $c_e$ and $c_t$ are the linear and torsional damping constants of the FREE. The best estimates of the stiffnesses $k_e$ and $k_t$ are empirically identifiable as studied in [41] under various static axial and torsional loading, assuming a linear spring relationship for the elastomer, using Eqs. (1) and (2). Our experimental measurements showed only small variations ($< 1\%$) in the values of $k_e$ and $k_t$ over a range of pressures, and therefore we considered them to be constant. The damping constants $c_e$ and $c_t$ are also computable by analyzing experimental data of rotational and axial vibration [41] in the form of a standard vibratory system [42]. Following the Newtonian relationship (shown in Fig. 3) between the force and moment components at pressure $P$, the final equations of motion are:

$$m_l \ddot{s} = F_l - k_e s - c_e \dot{s} + \pi r^2 P(1 - 2 \cot^2 \gamma) \tag{3}$$

$$I_l \ddot{\phi} = M_l - k_t \phi - c_t \dot{\phi} - 2\pi r^3 P \cot \gamma \tag{4}$$

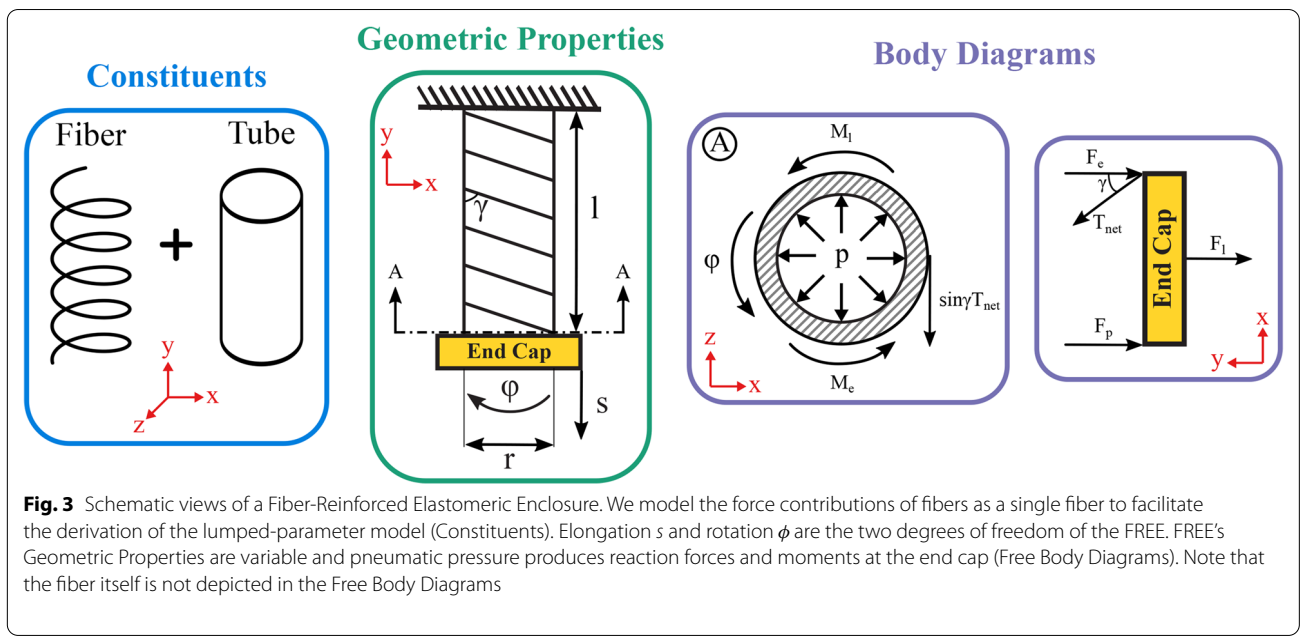

**Fig. 3** Schematic views of a Fiber-Reinforced Elastomeric Enclosure. We model the force contributions of fibers as a single fiber to facilitate the derivation of the lumped-parameter model (Constituents). Elongation $s$ and rotation $\phi$ are the two degrees of freedom of the FREE. FREE's Geometric Properties are variable and pneumatic pressure produces reaction forces and moments at the end cap (Free Body Diagrams). Note that the fiber itself is not depicted in the Free Body Diagrams

where $m_l$ and $I_l$ are the mass and the mass moment of inertia of the *end cap*, respectively. We consider the rotational mass moment of inertia of the FREE itself to be significantly less than any load inertia. The above equations present a lumped-parameter model of a FREE under certain simplifying assumptions based on the relationship between the applied forces, moments, and resulting reactions when fixed at one end and establish the relations between internal pressure and axial and rotational displacements. Note that Eqs. (3) and (4) are coupled. A variety of geometries and loading conditions can be considered by a designer to explore the dynamics of the system. The statically measured and predicted response of a FREE was studied in [39] under various input pressures and winding angles. In the following section, we examine the dynamic response of a FREE with a PID-controlled input pressure.

## Controller design

In this section, we go beyond analyzing the influence of design parameters and focus on developing a simple controller to study the dynamic motions of a single FREE. Previous works formulated based on the infinite number of degrees of freedom in soft-bodied actuators have proven the practicality of feedback control for soft robotic systems [15, 43, 44]. One challenge however is the lack of a unified framework for finding an appropriate controller for these soft robotic systems [45]. For the scope of this research, we seek to explore using a lumped-parameter model as a streamlined approach to developing a PD controller for FREEs. Also, we emphasize that a simple controller can be sufficient for a FREE to follow a particular trajectory and minimize oscillatory behavior. This makes soft actuators practical in many scenarios that require compliant and safe interaction with the environment. For instance, imagine twisting a knob using a robot arm with a FREE at the end-effector. FREEs are particularly capable of producing rotation (see Finite Element Model section) and a simple controller can provide an accurate and smooth motion. Our approach provides insight into controller development for FREEs and ultimately expands the understanding of soft robotic systems in which these actuators are used. Here, we elucidate controlling the rotational motion of a single FREE without external loading. We empirically show that using a simple PID (Proportional-plus-Integral-plus-Derivative) controller appropriately regulates the internal pressure $P$ (Eq. 5) to control the rotation of a FREE over time $t$. This simple controller is given by

$$P = K_p(\phi_d - \phi) - K_d\dot{\phi} + K_i \int (\phi_d - \phi)\, dt \qquad (5)$$

where $K_p$, $K_i$, and $K_d$ are the proportional, integral, and derivative gains, respectively, that drive the internal pressure of FREE to achieve the desired rotation angle $\phi_d$. Note that the axial motion of a FREE is generally not significant compared to its rotation; therefore, we neglected the elongation $s$ in developing this linear controller. Note that the desired angular velocity $\dot{\phi}_d$ is set to zero. Plugging the terms of Eq. (5) into Eq. (4) results in a model describing the controlled rotation angle $\phi$ as a function of time given by:

$$B\ddot{\phi} + C\dot{\phi} + E\phi + F \int \phi\, dt = \\ D\phi_d + F \int \phi_d\, dt + M_l \qquad (6)$$

where

$$\begin{cases} B = I_l \\ C = (-2\pi r^3 \cot \gamma)K_d + c_t \\ D = (-2\pi r^3 \cot \gamma)K_p \\ E = D + k_t \\ F = (-2\pi r^3 \cot \gamma)K_i \end{cases} \qquad (7)$$

Experimental determination of suitable state feedback gains that produce stable closed-loop pole locations for the system can be tedious. To facilitate the process, we employed the classical root-locus method to find the roots of the characteristic equation and initial values for each gain (i.e., $K_p$, $K_i$, and $K_d$). In order to apply the root-locus method, Eq. (6) must be linearized about an equilibrium state, in this case $s$, $\phi = 0$. Chapter 3 in [41] discusses the linearization process in detail using a Taylor series expansion about the equilibrium state. After linearization, the radius $r$ and winding angle $\gamma$ in Eq. (7) are replaced with the initial constant values R and Γ, respectively. In other words, the linearized equations of motion do not capture changes to the FREE's radius and winding angle during pressurization. Figure 4 depicts the block diagram of the PID control system, where the linearized equation of rotational motion is used. Note that we assumed perfect sensor feedback (i.e., $H(s) = 1$). To examine the impact of neglecting changes in radius and winding angle on overall response, Fig. 5 presents open-loop simulations results using both the linearized and nonlinear equations of motion. As the results demonstrate, $r$ and $\gamma$ only change 2% at maximum for the particular parameters considered. The linearized model overestimates the required input pressure by about 10%; however, the overall response of the system is approximately the same. The model based on the linearized equations of motion reaches the final rotation angle as fast as (within 0.2 sec) the model based on the fully nonlinear terms in Eq. (6). This indicates that the assumptions made in the linearized model do not affect the system's behavior at low pressures.

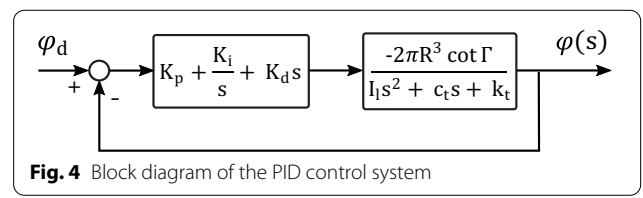

**Fig. 4** Block diagram of the PID control system

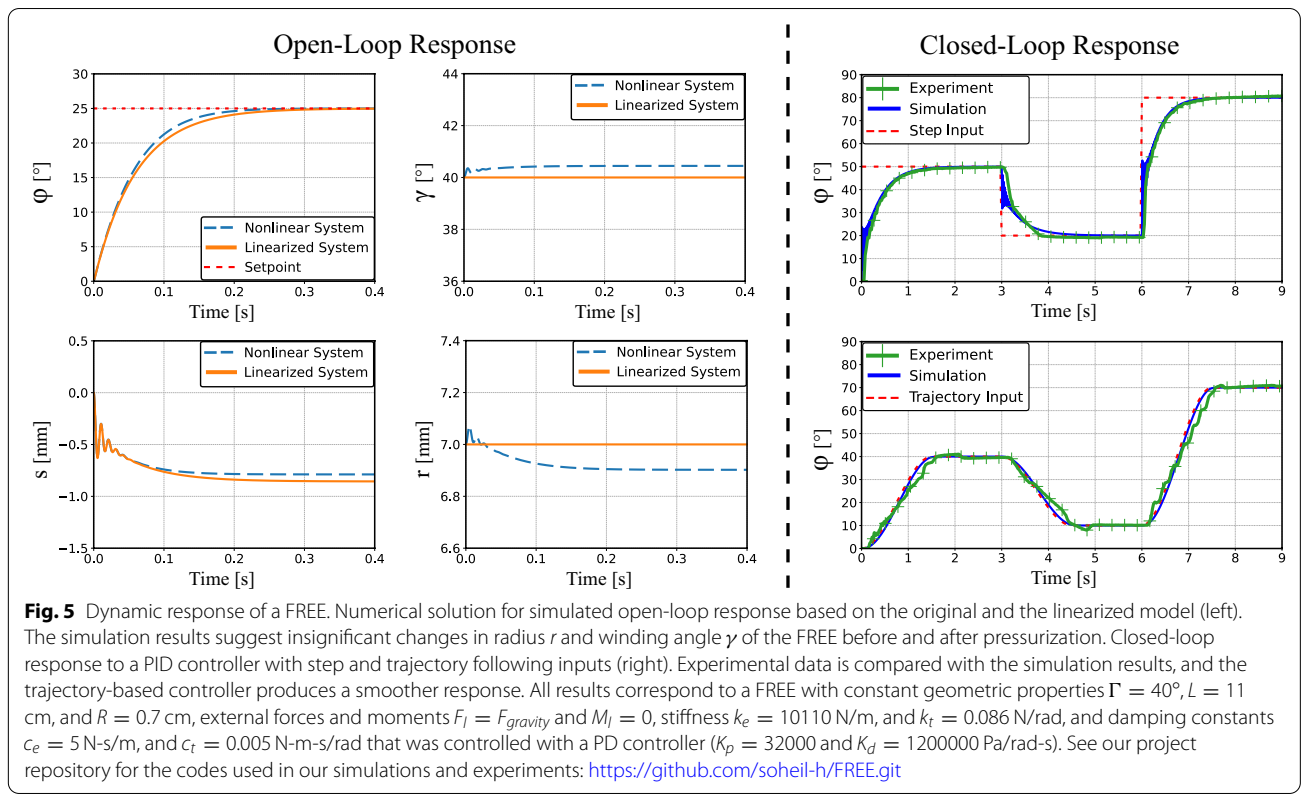

**Fig. 5** Dynamic response of a FREE. Numerical solution for simulated open-loop response based on the original and the linearized model (left). The simulation results suggest insignificant changes in radius $r$ and winding angle $\gamma$ of the FREE before and after pressurization. Closed-loop response to a PID controller with step and trajectory following inputs (right). Experimental data is compared with the simulation results, and the trajectory-based controller produces a smoother response. All results correspond to a FREE with constant geometric properties $\Gamma = 40°$, $L = 11$ cm, and $R = 0.7$ cm, external forces and moments $F_l = F_{gravity}$ and $M_l = 0$, stiffness $k_e = 10110$ N/m, and $k_t = 0.086$ N/rad, and damping constants $c_e = 5$ N-s/m, and $c_t = 0.005$ N-m-s/rad that was controlled with a PD controller ($K_p = 32000$ and $K_d = 1200000$ Pa/rad-s). See our project repository for the codes used in our simulations and experiments: https://github.com/soheil-h/FREE.git

**Experiment results—PID control**

We have proposed a model to predict the dynamic behavior of a FREE and shown that simplifying assumptions (i.e., constant radius and winding angle) adequately serve the purpose. In this section, we present experimental test results determined with an actual FREE and discuss the observed behavior as compared to simulation. We studied the closed-loop response with two types of control inputs. The first type delivers a step input every three seconds to the controller of positive 50°, negative 30°, and positive 60°. The second input is a cubic polynomial trajectory that generates a similar rotation angle variation by commanding desired rotation angles of first 40°, then 10°, and finally 70° while creating a smooth rotational motion. The equation for the commanded trajectory is:

$$\phi(t) = a_0 + a_1 t + a_2 t^2 + a_3 t^3 \qquad (8)$$

where $\phi$ is the rotation angle of the free end of the FREE, and $a_i$ (i= 0,1,2,3) are coefficients based on satisfying the constraints of the motion. Defining $t_f$ and $\phi_f$ as the desired duration of the maneuver and the goal angle of rotation, respectively, the initial and goal position of rotation are $\phi(0) = \phi_0$ and $\phi(t_f) = \phi_f$. Specifying the angular velocities at the beginning and the end of the trajectory as $\dot{\phi}(0) = 0$ and $\dot{\phi}(t_f) = 0$, provides two more constraints. By substituting these conditions into Eq. (8) and its time derivative and solving for the $a_i$, the following trajectory equation is obtained:

$$\phi(t) = \phi_0 + \frac{3}{t_f^2}(\phi_f - \phi_0) - \frac{2}{t_f^3}(\phi_f - \phi_0) \qquad (9)$$

Using Eq. (9) as the controller input generates maneuvers in which the actuator reaches the final position smoothly. We tested the system using both step and trajectory controller inputs and simulated the corresponding system responses using the dynamic model. We used a Bellofram type T-1500 pressure regulator to supply air, and an Adafruit BNO055 IMU (Inertial Measurement Unit) to measure the rotation of the FREE. To control the pressure regulator we used an Arduino microcontroller and acquired rotation data through I2C communication. The results of the simulations and experiments for step and trajectory inputs for a typical FREE are displayed in Fig. 5. Note that we experimentally measured the stiffness and damping constants of the FREE for simulation purposes. Comparing experimental and simulation results suggests that the lumped-parameter model fairly predicts the motion of the FREE with the PID controller. For the step input response, the RMSD (root-mean-square deviation) between simulation and experimental is 4% over the total duration of the maneuver. There are slight differences in the shape of curves, especially at the beginning of each step, due to 1) the sampling rate in the simulation being considerably higher than in data sets collected by the experiment and 2) the physical characteristics of the pressure regulator not being included in the simulation.

As was the case for the step input, the trajectory following experiment yielded results matching the simulation, although the relatively low speed of the Arduino control loop created a trajectory in the experiment that exhibited small oscillations relative to the commanded path. The RMSD between simulation and experimental results was 3% over the total duration of the maneuver.

One important insight obtained from the trajectory following simulation, besides its importance to actual applications, is the realization that high-frequency oscillations observed in the step input response are likely due to rapid changes in the desired angle of rotation. In the trajectory following maneuvers, the desired angle varies gradually and the oscillations are not present. Note however that the oscillations can likely be minimized by introducing derivative feedback. In general, a trajectory following control input is a better alternative, especially in multiple FREE modules, for creating smooth motions.

## Finite element model

The previous section presented a lumped-parameter model that predicts the behavior of FREEs and assists a designer in avoiding tedious and time-consuming build-and-test processes. Our proposed mathematical model is quite useful for obtaining an understanding of the dynamic behavior of a single FREE. However, establishing an idealized lumped-parameter model for multiple FREEs in a module is more challenging. Creating a matrix formulation for equations of motion of each FREE in a module that converges to a numerical solution is not guaranteed in particular parameter regimes. More significantly, formulating the equations governing the relationships between moments and forces at the end effector of a module is complicated and fraught with error.

Finite element analysis (FEA) has previously been used to study the effect of fiber orientations on FREE mechanics [30, 32]. There are a variety of parameters that affect such an actuator's response, including both geometric and material properties. We seek to study the choice of these parameters in designing FREEs in particular applications and in achieving desired overall response characteristics. In this section, we focus our attention on developing a finite element model of a FREE in single and module configurations (Fig. 6) to explore responses, to consider the impact of parameter variations, and to overcome the mathematical modeling difficulties inherent with multiple FREEs.

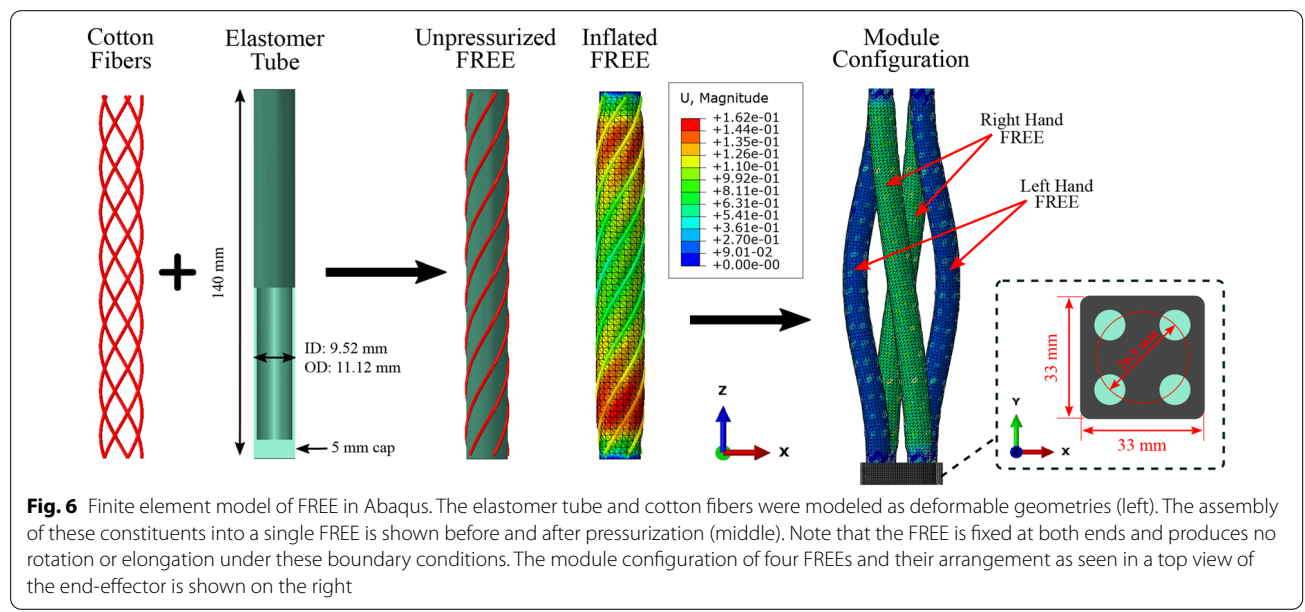

**Fig. 6** Finite element model of FREE in Abaqus. The elastomer tube and cotton fibers were modeled as deformable geometries (left). The assembly of these constituents into a single FREE is shown before and after pressurization (middle). Note that the FREE is fixed at both ends and produces no rotation or elongation under these boundary conditions. The module configuration of four FREEs and their arrangement as seen in a top view of the end-effector is shown on the right

## Model formulation

The extent of finite element analysis of soft pneumatic actuators in the past has been limited and only found to be useful for special cases of materials and geometries [46]. Generally, the research has been narrowly focused on the linear behavior of soft materials at small strains [47, 48]. Various constitutive models have been investigated to understand the hyperelastic behavior of soft materials such as the Mooney–Rivlin [49] and neo-Hookean [32]. These models are based on linear approximations of the strain invariants and also limited to small strains [50].

We built our modeling approach based on a previously published analysis of soft actuators [32] and extended the analysis to capture the nonlinear behavior of FREEs. The geometry of our model includes three regions: a deformable three-dimensional elastomeric tube, rigid end caps, and deformable fibers wound on the exterior of the tube (see Fig. 6). As discussed previously, pressurized FREE exhibits three distinct deformations: extension, expansion, and rotation. Rotation is typically the primary interest since other deformations generally do not contribute substantially to overall behavior based on results achieved in our analysis [41].

Similar to our approach in developing the lumped-parameter model, we made simplifying assumptions when developing the FEA model: fibers are modeled as perfectly connected to the elastomer tube, the effect of the rubber cement coating on the behavior of the FREEs has been neglected, and all fibers are assumed to initially match the prescribed winding angle. In this analysis, the elastomer geometry was modeled in the finite element analysis package Abaqus with 16, 830 second-order hybrid tetrahedral elements. Fibers were modeled with 1, 356 second-order truss elements, which only support axial tensile or compressive loads and not shear or bending. Contact and adhesion between elastomer and fibers were modeled with tied constraints. Desired variables of the FREE were carefully calculated by tracking a set of elastomer surface nodes as a function of pressurization. All models were run in Abaqus/Standard from Dassault Systèmes with the proximal cap face pinned and pressure applied to the entirety of elastomer interior surface.

## Numerical modeling

To investigate FREEs' behavior, we considered various constitutive models for the elastomer. These included a linear elastic and two different hyperelastic models: neo-Hookean and first-order Ogden. We first obtained the true stress-strain relationship for the latex elastomer through both uniaxial material testing using an Instron 5965 with a dogbone sample and through axial stretching of an elastomeric tube. The resulting linear curve fit provided a Young's Modulus (1.18 MPa) for the linear elastic model. We further checked that the data were in a reasonable range by calculating the elastic modulus corresponding to typical Shore A hardness numbers for latex ($35 \pm 5$). An elastic modulus of 1.18 MPa corresponds to a Shore hardness of 30.8 based on the formula given in [51], which is within the expected range.

The Ogden model represented by Eq. 10 is a polynomial [52, 53] that normally yields good results capturing the mechanics of soft materials. Eq. 10 presents the strain density energy for a first-order Ogden model, where $\lambda_i$ (i= 1,2, and 3) represent the deviatoric principal stretches, $\mu$ is the shear modulus, and $\alpha$ is a material constant.

$$\Psi = \frac{2\mu}{\alpha^2}(\bar{\lambda}_1^\alpha + \bar{\lambda}_2^\alpha + \bar{\lambda}_3^\alpha - 3) \tag{10}$$

Once the material and structural properties of the elastomer were determined, we conducted experiments on a simple latex tube without a fiber winding [45] and evaluated each constitutive model (i.e., linear, neo-Hookean, and Ogden). Finite element simulations of this experiment led us to discover that the neo-Hookean model predicts *a softer* response than the actual FREE at high pressures. The linear model was not found to be suitable either since it completely neglects the nonlinear behavior of the FREE. Therefore, we focused on the use of a first-order Ogden hyperelastic model in our analysis.

Employing an Ogden model requires the determination of the initial shear modulus $\mu$ as well as the material parameter $\alpha$ that characterizes nonlinearity. Based on the previously determined elastomer modulus, we chose to fix our initial shear modulus at 0.393 MPa (converted from $E = 1.18$ MPa with $\nu = 0.5$) and vary $\alpha$ through a sensitivity study. The error measurements between models and the experimental data showed that an Ogden model with $\alpha = 1.2$ to be the best representation of the behavior of a coated FREE (RMSD with $\alpha = 1.2$ was 50% less than with $\alpha = 0.8$ and 85% less than with $\alpha = 2$). Note that the use of other elastomer materials would likely yield other values of $\alpha$ but were not considered in the present work.

Fibers were modeled to have linear elastic material properties. Similar experiments were performed for cotton fibers to determine the tensile force-strain relationship. Load-strain data were acquired to provide the relationship between fiber strain and the product of modulus and cross-sectional area of the fiber (EA). This enabled the use of truss elements for cotton fibers in the finite element model. Fitting the linear region of the test results gave a value of structural stiffness equal to 644 N/$\epsilon$. As shown in [45], the exact value of EA is in fact not critical to the analysis due to the significantly greater stiffness of the cotton fibers relative to the latex elastomer (Fig. 7).

### Experiment—finite element model

We defined the finite element model formulation, in which linear elastic and hyperelastic Ogden models are used for the fiber and elastomer, respectively. Now, we present the model validation. This follows our previous work [45] where we used the model built from tensile and expansion data (elastomer $\mu = 0.393$ MPa and $\alpha = 1.2$, fiber $EA = 644$ N/$\epsilon$) to accurately predict FREE rotation. We chose rotation for validation as it is the principal deformation of interest for most FREEs.

Figure 6 displays curves characterizing the rotation per unit length ($\tau$) of FREEs with fiber winding angles of 20°, 40°, and 70° as predicted by our calibrated FEA model along with experimentally measured rotations from the actual FREEs. Overall, these results confirm that our model has strong predictive capabilities across various fiber angles over a range of pressurization. At higher pressures, we noticed discrepancies between the model and experiments for the 20° and 40° cases. The RMSD was 11.8% for the 20° and 9.5% for the 40° cases, while it was 11.8% for the 70° FREE. Potential reasons for these discrepancies include slight deviations in the experimental winding angle, the assumption of idealized tied connectivity between the elastomer and fibers, the effect of the rubber cement coating on experimental FREE structural properties, and slight variations that may exist between the Ogden material model and the material properties of the elastomer.

Comparing our results to prior work, the research by Connolly *et al.* [32] is the most relevant to our work. While the material properties and our modeling approach differ from those used in [32], we were able to use the same modeling framework to develop our geometry and observed similar qualitative trends. Connolly et al. [32] used second-order, one-dimensional beam elements to model Kevlar fibers in their actuators, which is appropriate because the bending stiffness of the Kevlar fibers contributes to the overall response. In our case, the cotton fibers do not support bending in response to the internal pressure and axial loads. Thus, we chose second-order truss elements for the fibers, which enabled us to simulate FREE's experimentally observed buckling behavior [45]. Also, Connolly et al. [32] used a neo-Hookean model for the elastomer (silicone) and reported convergence issues at large fiber angles ($\geq 80°$). To avoid these issues, they conducted a deformation-driven analysis by simulating the internal pressure through thermal expansion. In contrast, we primarily used an Ogden model. Our implementation of an Ogden hyperelastic model enabled a better fit with experiments [45] results, and we did not observe any convergence issues. In summary, our approach to modeling fiber-reinforced soft actuators allowed us to capture the buckling behavior of these actuators and resolve the convergence issues experienced using a neo-Hookean model.

### Significance of winding angle

One of the principal motivations of creating the finite element model was to avoid directly measuring system parameters (e.g., damping constant and stiffness) of each FREE. Otherwise, the designer has to determine these parameters by running experiments on every single actuator in order to use the lumped parameter model. The finite element model on the other hand ideally only requires one-time calibration based on the properties of constituent materials. Additionally, our previous parameter studies [45] showed that only variations in elastomer material properties significantly influence FREE behavior because the stiffness of the fiber is relatively so much higher. Hence, the designer is able to tune the model by only measuring the material properties of the elastomer.

So the question then arises, what overall contribution do the fibers make to the behavior of FREEs? To answer this question, we explored the effects of variations in fiber winding angle on FREEs' behavior. Figure 8 shows the behavior of FREEs with winding angles $\Gamma = 10°$ to $\Gamma = 80°$ in terms of rotation per unit length $\tau$, axial extension $\lambda$ (deformed length/initial length), and force and moment generation. This provides a road map for the designer to pick an actuator that best suits the desired application.

For motion analysis, we defined a pinned constraint at only one end of the actuator in the finite element model and measured the displacements of the free end.

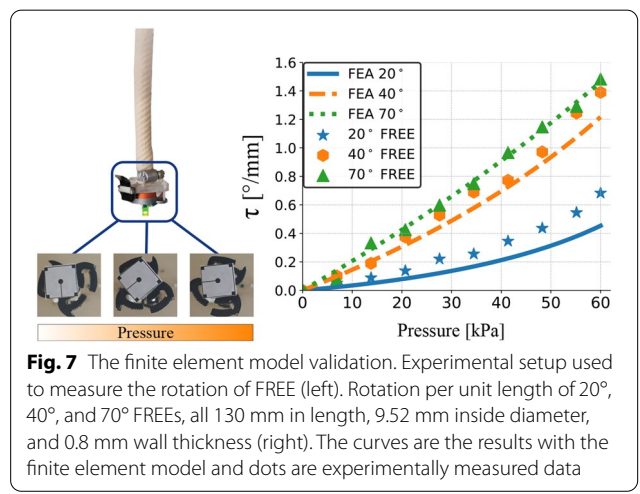

**Fig. 7** The finite element model validation. Experimental setup used to measure the rotation of FREE (left). Rotation per unit length of 20°, 40°, and 70° FREEs, all 130 mm in length, 9.52 mm inside diameter, and 0.8 mm wall thickness (right). The curves are the results with the finite element model and dots are experimentally measured data

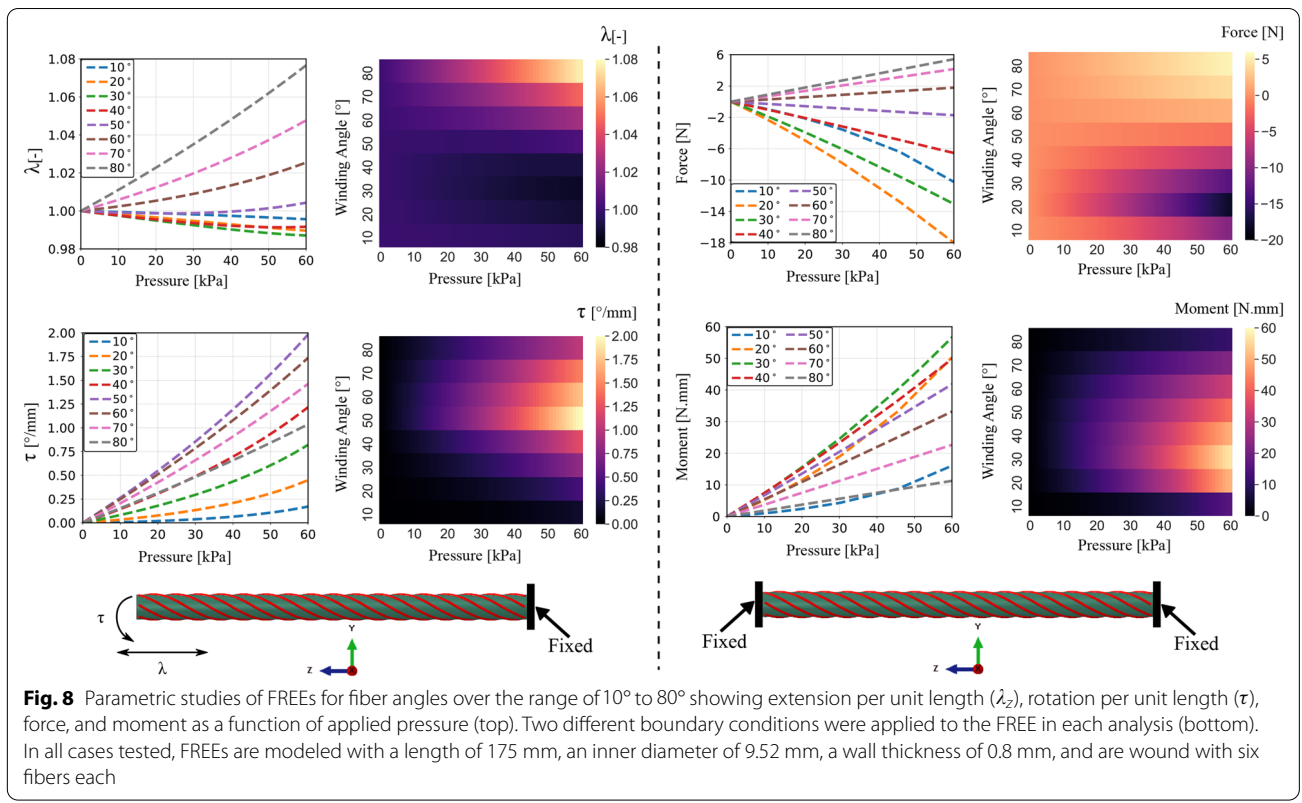

**Fig. 8** Parametric studies of FREEs for fiber angles over the range of 10° to 80° showing extension per unit length ($\lambda_z$), rotation per unit length ($\tau$), force, and moment as a function of applied pressure (top). Two different boundary conditions were applied to the FREE in each analysis (bottom). In all cases tested, FREEs are modeled with a length of 175 mm, an inner diameter of 9.52 mm, a wall thickness of 0.8 mm, and are wound with six fibers each

In practice, soft actuators that generate large deformations are practical from a design perspective since they provide a reasonable range of motion. Our results indicate that as pressure increases, FREEs with winding angles between 50° and 70° produce the greatest rotations, while winding angles between 10° and 20° lead to considerably smaller amounts of rotation (25% less than a FREE with a 50° winding angle). For extension, 70°-to-80° FREEs are most capable of generating axial motions (∼14 mm), while FREEs with winding angles between 10° and 50° exhibit very limit axial deformation. As shown in Fig. 8, FREEs are best as rotational actuators rather than producing axial motions, particularly as compared with alternatives such as McKibben actuator [28]. In order to measure FREEs' force and moment generation capabilities, we modeled the actuator with both ends pinned (i.e., no rotational or axial displacements). Our analysis reveals that 10°-to-30° FREEs produce the largest force magnitudes. Conversely, the axial force generated by 50°-to-60° FREEs is negligible compared to FREEs with other winding angles. We also found that 20°-to-50° FREEs are more capable of producing moments than those that have greater or lesser winding angles. Note that we chose

10° increments between different winding angles due to the manufacturing constraints in aligning fibers.

Looking at the results of the motion analysis, one might expect that the FREE that exhibits the largest deformations should produce the largest forces and moments. This was not observed however due to the differences in boundary conditions. In the displacement analysis ($\lambda$ and $\tau$), the FREE is only fixed at one end, while in the force and moment analysis the FREE is fixed at both ends. Since a FREE is an underactuated system, any type of constraint significantly affects the behavior. We have thus commented on the trends across winding angles when interpreting the results of each analysis individually.

Our results suggest that the winding angle primarily determines the behavior of a FREE because fibers are inherently much stiffer than the elastomer. Some winding angles exhibit particular behaviors (e.g., a 50° FREE produces nearly a pure rotation) and it is the designer's choice to pick a specific winding angle for their application. Overall, the finite element model is particularly advantageous for applications that require specific design considerations before manufacturing. This enables convenient tuning of various parameters, such as winding angle and material properties of the elastomer, and probing the particular behavior of this type of actuator.

## FREE module

In the previous sections, we discussed modeling the behavior of a single FREE. Here, we take a step forward in using finite element analysis and study multiple FREEs in a module (Fig. 6). Studying arrangements of FREEs in a module is particularly beneficial because, in contrast to a single FREE, the module can generate maneuvers in a three-dimensional space. Hence, FREE modules have the potential to become practical elements of a complete soft robotic arm.

The finite element model allows us to explore various FREE winding angles, geometric arrangements, and actuation combinations in a module before conducting experiments. This section presents the results of finite element analyses of the motion of modules consisting of FREEs with 30° and 60° winding angles. In exploring module configurations, we realized that the winding direction of the fibers is an important factor in designing a FREE module. Here, we use L and R notations for

FREEs with clockwise and counterclockwise winding directions, respectively. For the modules considered here, each module consists of four FREEs, one pair of each winding direction (two L and two R FREEs, denoted as an LR module) with each L or R FREE positioned diagonally across from one another. Each FREE can be actuated individually within the module.

To study all possible actuation combinations, we considered a total of 5 cases. Case 1 corresponds to all four actuators being pressurized—one variation. Case 2 corresponds to each of two diagonally opposed actuators being pressurized—two variations: one for each diagonal, one consisting of two L actuators, and one of two R actuators. Case 3 pressurizes one FREE at a time—four variations. Case 4 pressurizes two adjacent FREEs—four variations: each consisting of one L and one R actuator. And Case 5 consists of pressurizing three actuators—four variations. The five cases, with a total of 15 variations, create the basis for determining the workspace of the module as depicted in Figs. 9 and 10. In all cases tested, FREEs—arranged in the square configuration shown in Fig. 6—are modeled with a length of 175 mm, an inner diameter of 9.52 mm, a wall thickness of 0.8 mm, and are wound with six fibers each. Figure 6 shows a finite element model of a FREE module in Abaqus.

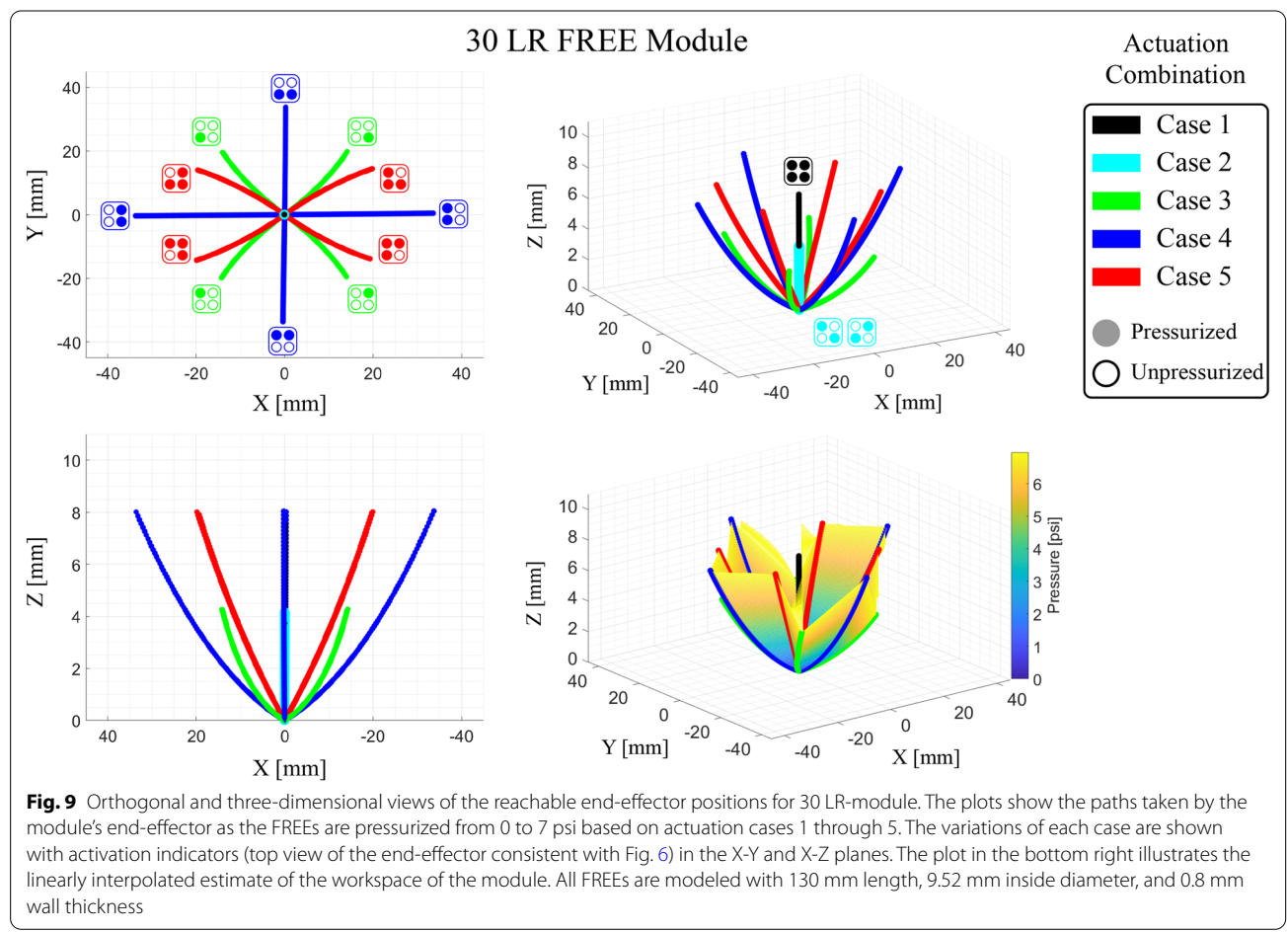

**Fig. 9** Orthogonal and three-dimensional views of the reachable end-effector positions for 30 LR-module. The plots show the paths taken by the module's end-effector as the FREEs are pressurized from 0 to 7 psi based on actuation cases 1 through 5. The variations of each case are shown with activation indicators (top view of the end-effector consistent with Fig. 6) in the X-Y and X-Z planes. The plot in the bottom right illustrates the linearly interpolated estimate of the workspace of the module. All FREEs are modeled with 130 mm length, 9.52 mm inside diameter, and 0.8 mm wall thickness

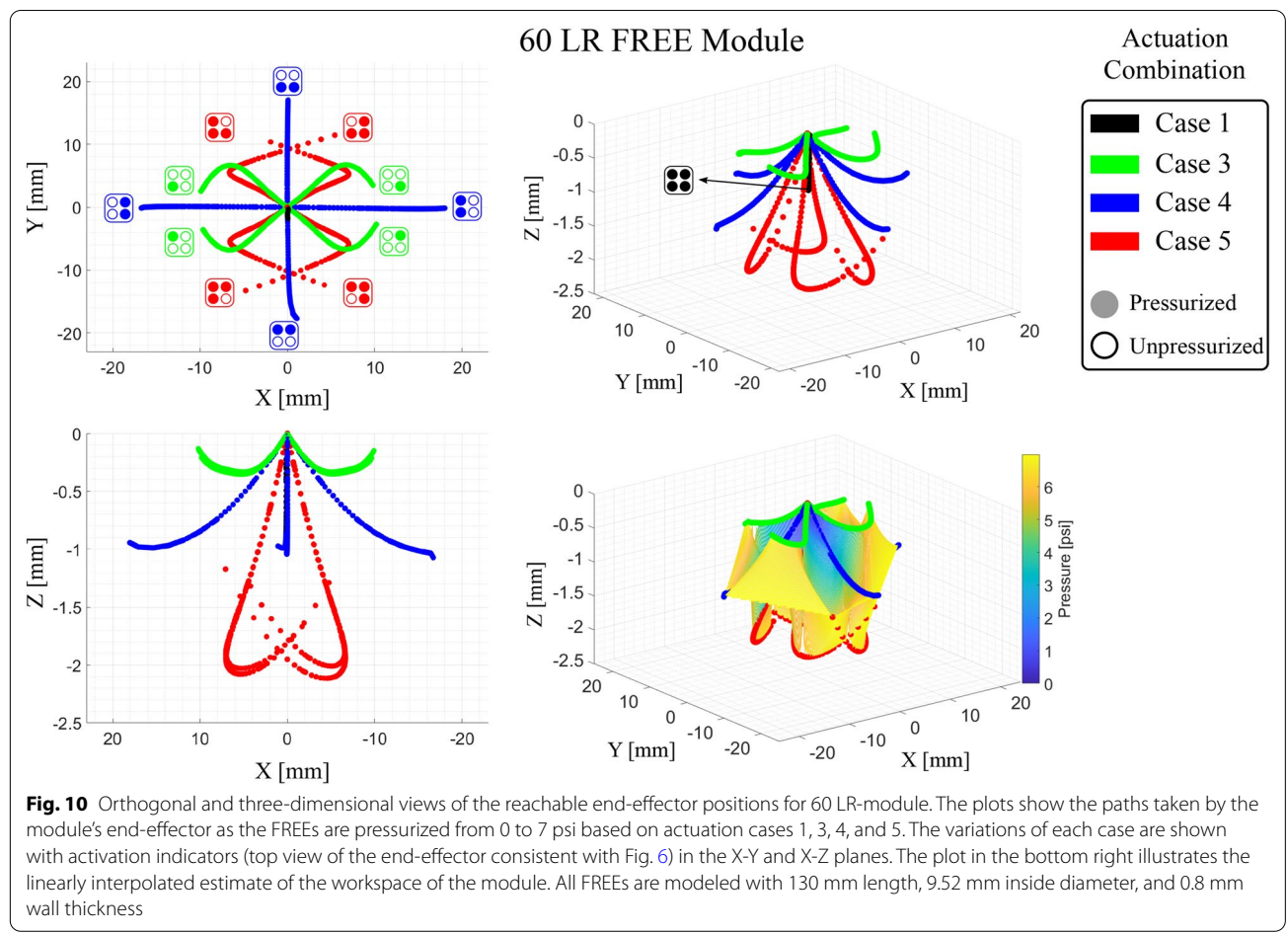

**Fig. 10** Orthogonal and three-dimensional views of the reachable end-effector positions for 60 LR-module. The plots show the paths taken by the module's end-effector as the FREEs are pressurized from 0 to 7 psi based on actuation cases 1, 3, 4, and 5. The variations of each case are shown with activation indicators (top view of the end-effector consistent with Fig. 6) in the X-Y and X-Z planes. The plot in the bottom right illustrates the linearly interpolated estimate of the workspace of the module. All FREEs are modeled with 130 mm length, 9.52 mm inside diameter, and 0.8 mm wall thickness

When interpreting the plots in Figs. 9 and 10, remember that the variations of each actuation case are shown with a unique activation indicator in the X-Y and X-Z planes. The activation indicator shows the module's end-effector from a top view (see Fig. 6): empty circles and filled circles correspond to unpressurized and pressurized FREEs, respectively. For instance, Case 4 has four different variations depending on which two pairs of FREEs are pressurized in the module. One of the variations is shown with the blue activation indicator positioned at (20,0) and shows that the end-effector travels from (0, 0) to (20, 0) when two FREEs on the left half of the module are pressurized together.

To identify the relationship between module pressurization and displacement, we ran the model for two LR-modules, one consisting of FREEs with a 30° winding angle and one consisting of FREEs with a 60° winding angle, and explored the reachable workspace of each module for the five actuation cases. Figures 9 and 10 illustrate the position of the end-effector of the 30° and 60° modules in space, respectively. The pressures within the pressurized FREEs were varied from 0 to 7 psi based on each actuation case (as described below, results suggest that cases 1, 2, and 4 are the most fundamentally useful in creating particular motions).

As the curves in Fig. 8 indicate, winding angles of 30° and 60° generate markedly different deformations in a single FREE: 1) a pressurized 30° FREE contracts while a 60° FREE elongates, and 2) a FREE with 60° winding angle rotates more than two times as much as a 30° FREE. These differences lead to different end-effector paths for 30° (Fig. 9) and 60° (Fig. 10) LR-modules, although the characteristic shapes of the case 1, 2, and 4 paths are similar in both figures.

Based on the results in Figs. 9 and 10, we found that cases 1, 2, and 4 are the most fundamentally useful in creating axial, torsional, and bending motions. Pressurizing all of the FREEs (Case 1) produces pure elongation for the 60° LR-module and pure contraction for the 30° LR-module (black paths in Figs. 9 and 10). Case 2 produces rotation without bending in an LR-module (cyan paths in Figs. 9 and 10). Note that the paths in both of these cases are overlapping vertical lines. It should be mentioned that we eliminated the results for Case 2 for the 60° LR-module from Fig. 9 because this case produces pure rotation (i.e., no elongation). Note that in our simulations with cases 1 and 2, we observed buckling in the response of the FREEs in a 60° LR-module [45]. Buckling arose for these cases as a result of large deformations—contraction in Case 1 and twist in Case 2.

Pressurizing two adjacent FREEs (Case 4) causes the module to deform in pure bending (blue paths in Figs. 9 and 10). The orthogonal views of the workspace show the projections of the paths of each case in the X-Y and X-Z planes. cases 1 and 2 generate one-dimensional lines, while cases 3, 4, and 5 result in paths generated through bending and twisting of the module. One notable observation from the results is the distinction between the directions of the end-effector paths for the two modules. The free end of the module extends upward for the 30° LR-module, while it extends downward for the 60° one. Additionally, the 60° LR-module displays more variations in the paths of motion for actuation cases 3 and 5, resulting in curved projections in the X-Y plane that extend approximately 10 mm from the origin. The pure bending produced by Case 4 results in straight-line projections of the paths in the X-Y plane that extend approximately 36 mm and 20 mm from the origin for the 30° and 60° LR-modules, respectively. To approximate the entire shape of the workspace, we ran the model for actuation cases 3 through 5 to determine the lines of reachable points shown in Figs. 9 and 10 and linearly interpolated between them to estimate the workspace boundaries. In the analysis described here, we only studied reachable points in space; the end-effector has its own unique set of orientations in each case.

The three-dimensional plots in Figs. 9 and 10 show the points reached as the pressure within the FREEs is increased from 0 to 7 psi. As expected, the set of reachable points forms a concave-shaped workspace for the 30° LR-module. The shape of the workspace is more complex for the 60° LR-module. Comparing these workspace shapes, we found that a 30 LR module is capable of reaching above the neutral (unpressurized) position of the end-effector, while a 60 LR module's reach is mostly below the neutral position. Additionally, the end-effector of a 30 LR module travels larger distances from the neutral position: almost twice as far in the X-Y direction and three times as far in the Z direction.

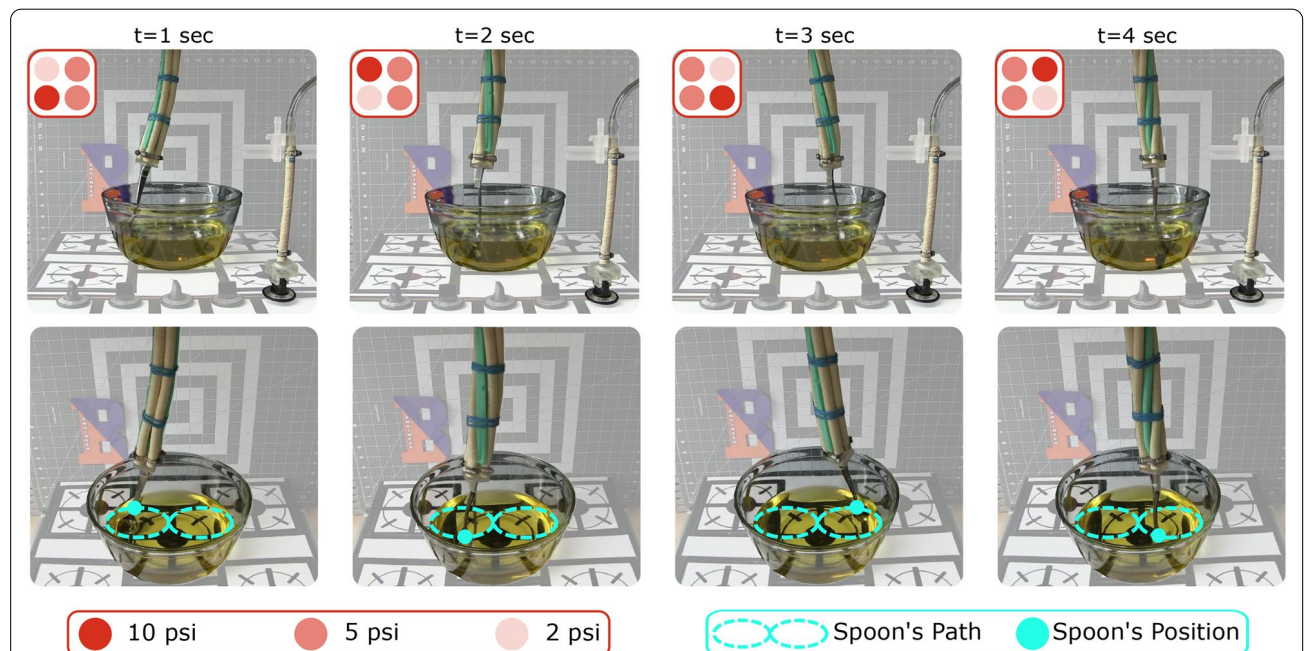

**Fig. 11** Cooking demonstration using FREEs. A single 50° FREE uses our proposed model-based PD controller (see Controller Design) to control the cooking temperature, and a 60 LR module uses a combination of cases 3 and 4 (see Fig. 10) to blend the ingredients of the bowl in a "figure eight" shape (cyan path). See the video of our experiment here: https://youtu.be/vwmvWCrHNQM

## Experimental demonstration

In this section, we present the experimental demonstration of a FREE's real-life application (Fig. 11). We show how insights from both modeling approaches can be used in a cooking scenario, where a single FREE controls the cooking temperature and a FREE module blends the ingredients in a bowl (Additional file 1). The single FREE uses our proposed model-based PD controller (Controller Design) to set the cooktop range knob at the desired temperature. Underneath the cooktop (not shown in Fig. 11) an IMU is attached to the range knob and feeds back the rotation angle to the controller. Our parametric analysis (Fig. 8) allowed us to find an appropriate winding angle for the single FREE: 50°. This winding angle is particularly useful since it produces a large rotation with minimum extension. For the stirring task, we replicated the widely embraced "figure eight" whisking technique using a 60 LR module. We employed our workspace study (FREE Module) and found that by combining the actuation combination of cases 3 and 4 (Fig. 10), a 60 LR module can create a "figure eight" whisking shape (cyan path in Fig. 11). In practice, we sinusoidally varied the pressure (2 psi to 10 psi) of each individual FREE within the module. With the sinusoidal input—over a specific period of time—the module is actuated with both case 3 and case 4: one FREE has the highest pressure (10 psi), one FREE has the lowest pressure (2 psi), and two FREEs have the same mid-range pressure (5 psi). The combination of both cases allows the generation of bending and twisting motions. This pattern interchanges every second among all FREEs of the module as shown in Fig. 11.

Although FREEs are still far from functioning as dexterous soft robotic arms, they present potential applications in individual and module configurations. This experiment not only demonstrated the static and dynamic motions of a FREE studied in our simulations but also provided us with further insights into FREEs' behavior under loading. In the future, we plan to continue to explore FREEs' payload capacity when creating motions.

## Conclusion

We formulated and analyzed two complementary modeling approaches for a particular type of soft robotic actuator, Fiber-Reinforced Elastomeric Enclosures (FREEs). Our principal insight was that even though both models effectively predict the behavior of FREEs, each model does not comprehensively represent the behavior of the entire system. Here, we developed both a dynamic lumped-parameter model and a finite element model in an attempt to understand the practicability of FREEs for use in a soft robotic arm.

We first introduced a lumped-parameter model based on the relationship between input pressure, internal and external loads, and geometric properties to predict the dynamic behavior of a single FREE. Our simulation analysis suggests that a FREE's winding angle and radius typically do not change more than 2% at low pressures. Based on this assumption, we tested the model with a PID controller to determine the response of the FREE to step and trajectory-following inputs. Both inputs generated responses in which the FREE successfully reached the desired rotation angle, demonstrating that a simple model is capable of predicting the motion of FREEs with less than 4% error. Extending the fundamental elements of this mathematical model to a module of multiple FREEs operating in a three-dimensional space, however, presents particular challenges.

We then developed a finite element model for single and multiple FREE configurations. We formulated the model using an Ogden hyperelastic material model for the elastomer and modeled the fibers as linearly elastic with 1D truss elements. The model was examined and validated by replicating experimental trials of FREEs' rotation. Our parametric studies indicate that FREEs with winding angles of 50°, 80°, 20°, and 30° demonstrate the greatest rotation, elongation, force, and moment, respectively. Using the finite element model, we were able to examine the impact of nonlinear FREE material properties as well as estimate the workspace of multiple FREEs in a module. Overall, we found the finite element model to be an effective tool and a step toward understanding the design space of FREEs in the module configuration.

Both of our modeling approaches have their shortcomings. The simplified lumped-parameter model neglects the nonlinearity of the system (i.e., assumes constant radius and winding angle), and we recognize the inherent inaccuracy in predicting behavior at high pressures. Also, measuring system parameters (i.e., damping constant and stiffness) for each FREE under consideration is required for the lumped-parameter model, and this could be a time-consuming process. The finite element model ignores manufacturing inconsistencies associated with the adhesive bonding fibers and the elastomer as well as the FREE outer coating. Moreover, the finite element model can be computationally expensive requiring long execution times, particularly for large loads and strains.

In summary, both models have the potential to significantly assist in the design process and to be expanded upon for studies of similar soft actuators. In future work, we plan to combine the insights from the workspace estimation for FREE modules with the trajectory following controller that we applied to a single FREE. This will potentially allow the end-effector of FREE modules to follow a specific path in space and reduce oscillatory behavior. Additionally, a more sophisticated hyperelastic material law could be developed to capture the viscoelastic effects of the nonlinear elastomer that we observed during experiments. Finally, our results suggest that FREE modules have a limited range of bending deformations and are more effectively used for generating rotational motions. For our future research, we plan to consider using other types of pneumatic actuators, such as McKibben actuators, combined with FREEs to create a complete, multifunctional soft robotic arm.

### Abbreviations
FREEs:: Fiber Reinforced Elastomeric Enclosures; PID:: Proportional-plus-Integral-plus-Derivative; IMU:: Inertial Measurement Unit; RMSD:: Root-Mean-Square Deviation; FEA:: Finite Element Analysis; L:: Clockwise Winding angle; R:: Counterclockwise Winding angle.

## Supplementary Information
The online version contains supplementary material available at https://doi.org/10.1186/s40648-022-00225-9.

---

**Additional file 1** FREEs demonstrate a cooking task in the real world.

---

### Acknowledgements
The authors would like to gratefully acknowledge the collaborative support of Prof. C. David Remy (University of Stuttgart), Dan Bruder (Harvard University), and Audrey Sedal (Toyota Technical Institute). They provided the impetus for our work as well as many thoughtful comments and refinements of our ideas. The authors also acknowledge the valuable insights provided by Prof. Christine Buffinton and Brielle H. Cenci of Bucknell University who contributed to the finite element analysis portion of the project.

### Authors' information
S. Habibian is a graduate student, S. Bae and J. Shin are undergraduate students in Mechanical Engineering. K. W. Buffinton and B. B. Wheatley are professors of Mechanical Engineering.

### Funding
Toyota Research Institute ("TRI") provided funds to assist the authors with their research but this article solely reflects the opinions and conclusions of its authors and not TRI or any other Toyota entity.

### Availability of data and materials
The data and source code used to support the findings of this study are available at https://github.com/soheil-h/FREE.

## Declarations

### Competing interests
The authors declare that they have no competing interests.

### Author details
[1]Was with Department of Mechanical Engineering, Bucknell University, Lewisburg, PA, USA. [2]Department of Mechanical Engineering, Virginia Tech, Blacksburg, VA, USA. [3]Department of Mechanical Engineering, Bucknell University, Lewisburg, PA, USA.

## Publisher's Note

Springer Nature remains neutral with regard to jurisdictional claims in published maps and institutional affiliations.